\documentclass[runningheads]{llncs}
\usepackage[T1]{fontenc}
\usepackage{graphicx}
\usepackage{comment}

\usepackage{siunitx}
\usepackage{comment}

\usepackage{graphicx}
\usepackage[normalem]{ulem}
\useunder{\uline}{\ul}{}
\usepackage{multirow}
\usepackage{adjustbox}
\usepackage{caption,subcaption,graphicx}
\usepackage{amsfonts}
\usepackage{booktabs}
\usepackage{float}
\usepackage{xcolor}
\usepackage{hyperref}
\usepackage{bbding}

\begin{document}

\title{A Multimodal Fusion Model Leveraging MLP Mixer and Handcrafted Features-based Deep Learning Networks for Facial Palsy Detection}
\titlerunning{A Multimodal Fusion Model for Facial Palsy Detection}
%

%


\author{Heng Yim Nicole Oo\inst{1}\orcidID{0009-0007-3504-5320} \and \\
Min Hun Lee\inst{1}\Envelope\thanks{Corresponding Author\\ Code: 
\textcolor{blue}{\href{https://github.com/harcs/MultimodalFusionAIFacialPalsy}{https://github.com/harcs/MultimodalFusionAIFacialPalsy}}}\orcidID{0000-0003-3506-8972}  \and \\
Jeong Hoon Lim\inst{2}\orcidID{0000-0002-4266-8709}}
\authorrunning{H.Y.N Oo et al.}
%
\institute{Singapore Management University, Singapore, Singapore \and
National University of Singapore, Singapore\\
\email{nicoleoo.2021@scis.smu.edu.sg, mhlee@smu.edu.sg, mdcljh@nus.edu.sg}
}

%

\maketitle              
\begin{abstract}
Algorithmic detection of facial palsy offers the potential to improve current practices, which usually involve labor-intensive and subjective assessments by clinicians. In this paper, we present a multimodal fusion-based deep learning model that utilizes an MLP mixer-based model to process unstructured data (i.e. RGB images or images with facial line segments) and a feed-forward neural network to process structured data (i.e. facial landmark coordinates, features of facial expressions, or handcrafted features) for detecting facial palsy. We then contribute to a study to analyze the effect of different data modalities and the benefits of a multimodal fusion-based approach using videos of 20 facial palsy patients and 20 healthy subjects. Our multimodal fusion model achieved 96.00 F1, which is significantly higher than the feed-forward neural network trained on handcrafted features alone (82.80 F1) and an MLP mixer-based model trained on raw RGB images (89.00 F1).


\end{abstract}

\section{Introduction}
Patients with facial palsy usually suffer from symptoms, such as drooping mouth and eyebrows, difficulty in closing an eyelid, and drooling. Facial palsy can have serious consequences on patients, such as diminished feeding function, psychological distress, and social withdrawal \cite{kosins2007}. For the diagnosis of facial palsy, clinicians usually perform observation-based physical examinations \cite{heckmann2019diagnosis,axioms12121091}. 
However, it is challenging to quantify symptom intensity and variation, track changes in these symptoms between visits for an individual patient, and compare differences in symptoms across different patients \cite{hsu2018deep}. 

To address this challenge, researchers have explored various algorithmic approaches to detect facial palsy \cite{hsu2018deep,Ngo2016QuantitativeAO,kim2015smartphone,Wang2014AutomaticRO}. These approaches broadly fall into two categories: 1) those employing machine learning models with manual feature extraction and 2) those that leverage deep learning-based models.

In this work, we investigate the performance of diverse data modalities for detecting facial palsy, including RGB images, facial landmark coordinates, and handcrafted feature vectors while analyzing the impact of utilizing different deep learning approaches. Specifically, we explore the benefits of using attention-based image models, which divide the input images into patches, for unstructured image data compared to traditional CNN-based models, such as ResNet50 \cite{he2016deep}. Furthermore, we present a multimodal fusion-based deep learning approach to explore the benefits of integrating structured and unstructured data modalities to enhance detection performance. By leveraging the YouTube Facial Palsy (YFP) dataset \cite{hsu2018deep} and the CK+ emotion dataset \cite{kanade2000comprehensive,lucey2010extended}, we provide a comprehensive evaluation of deep learning models using single and multimodal data.

Our evaluation reveals that single-modality models, such as a feed-forward neural network using handcrafted features, achieved a strong F1 score of 82.80, while an MLP Mixer-based model \cite{tolstikhin2024} utilizing RGB images reached an even higher F1 score of 89.00. Furthermore, the integration of RGB images and handcrafted features in our proposed multimodal fusion-based model led to a significant performance improvement, achieving an F1 score of 96.00. These findings demonstrate the advantages of combining diverse data modalities to enhance the accuracy and robustness of facial palsy detection.

Our primary contribution lies in assessing variations in deep-learning model performance across diverse data modalities (i.e. facial landmark coordinates, raw RGB images, etc), and demonstrating the benefit of integrating handcrafted manual features alongside image-based modalities to improve model performance in diagnosing facial palsy. This study establishes a benchmark for future research in facial palsy diagnosis, providing valuable insights into the contributions of individual modalities and the potential of multimodal fusion approaches.

\section{Related Work} \label{section:related}

\subsubsection{Manual Feature-based Approaches}
Manual feature extraction approaches have been extensively explored in facial palsy diagnostics research. These approaches typically fall into one of these three categories: Quantitative Analysis of Movement, 3D Scanning and Photogrammetry-Based Methods, and Hybrid and Automated Manual Feature Extraction.

Demeco et al. \cite{DEMECO2021102485} combined surface electromyography (sEMG) and kinematic analysis to assess muscle function, achieving high intra-subject reliability (ICC = 0.94) and interrater reliability (ICC = 0.91). Similarly, 
He et al. \cite{heshusoraghan2009} used Local Binary Patterns (LBP) to analyse motion in biomedical videos, measuring symmetry via resistor-average distance (RAD) between LBP features extracted from both sides of the face.

Meitjs et al. \cite{meintjs2003} introduced stereo-photogrammetry for non-intrusive, precise 3D facial measurements, originally designed to measure the facial dysmorphology in children with fetal alcohol syndrome, but adaptable for facial palsy diagnostics. 
Petrides et al. \cite{PETRIDES2021605} reviewed 3D scanning systems, including stereophotogrammetry and structured-light scanning for soft-tissue facial assessment.

The method by Kim et al. \cite{kimjunsikjeong2022} uses scale matching, global registration, and iterative closest point (ICP) techniques to align facial data. Symmetry is analyzed through distance, angular symmetry, and landmark movement on both sides of the face. Parra-Dominguez et al. \cite{diagnostics12071528} proposed a set of handcrafted features based on distance and angular symmetry computed from facial landmarks, achieved an accuracy of 97\% and precision of 94\% on the training dataset. 

For the manual feature-based approach, we follow the methodology outlined by \cite{diagnostics12071528} for the computation of handcrafted asymmetry measures due to their straightforward implementation and good results. For evaluation, the previous work \cite{diagnostics12071528} selects 20 samples per subject for both training and testing with k-fold cross-validation. To ensure a more rigorous evaluation, we increase the dataset size, selecting 40-50 samples per patient for training and testing, and evaluate model performance with Leave-One-Person-Out (LOPO) cross-validation.

\subsubsection{Deep Learning-based Approaches}
Hsu et al. \cite{hsu2018deep} proposed the deep-learning based approach to detect facial palsy. It comprises three component networks: the first detects the subject’s face, the second locates facial key points and line segments on the detected face, and the last identifies local palsy regions. Storey et al. \cite{storey8811497} introduced 3DPalsyNet, an end-to-end framework using a 3D CNN architecture with a ResNet backbone for mouth motion recognition and facial palsy grading. Arora et al. \cite{arora13080200} proposes a multi-model framework using an MTCNN for classifying facial palsy.

Delannoy-Ward et al. \cite{delannoyward2010} introduced a method that combines images of the left and right sides of a subject's face performing different expressions to create a new asymmetrical expression. Abayomi et al. \cite{abayomi10080978} introduces a novel data augmentation technique called Voronoi Decomposition-Based Random Region Erasing (VDRRE) to address challenges in deep learning-based facial palsy detection, including data scarcity and class imbalance.

Both Abayomi et al. \cite{abayomi10080978} and Yaotome et al. \cite{yaotome2019} leverage generative adversarial networks (GANs) to augment training datasets using diverse face images. Pourebadi et al. \cite{pourebadi2022} proposed a novel mathematical model to simulate the dynamics of facial expressions, enabling the creation of realistic synthetic faces for static image modeling and animations.

Although numerous approaches have been proposed for identifying and diagnosing facial palsy \cite{axioms12121091}, most researchers use private, in-house clinical data. A significant challenge in the field is the absence of a standardized framework for benchmarking, making it difficult to systematically evaluate and compare different approaches.

To address this, we utilized the publicly available YFP dataset \cite{hsu2018deep} and the CK+ emotion dataset \cite{kanade2000comprehensive,lucey2010extended} to conduct a comprehensive evaluation of deep learning models \cite{hsu2018deep,axioms12121091,oo2024exploring} using single and multiple data modalities. In addition, in contrast to previous studies \cite{hsu2018deep,axioms12121091,oo2024exploring}, our work further investigates the effectiveness of patch-based and transformer-based architectures, including MLP Mixer and FasterViT with image-based data modalities.

\section{Methods} \label{section:methods}
In this section, we describe how we process RGB images to obtain five independent data modalities and present the overall architectures of our single-modality and early/late fusion models with unstructured and structured modalities.
\begin{figure*}[htp]
\centering
\includegraphics[width=1.0\linewidth]{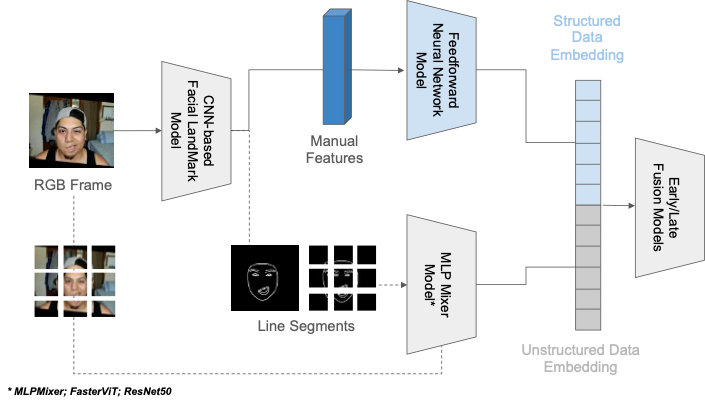}
\caption{Our early fusion model integrates any structured data embedding from a feedforward neural network with any unstructured data embedding from any image-based model to detect a patient with facial palsy}\label{fig:model_architecture}
\end{figure*}

\subsection{Data Processing \& Modalities}\label{sect:data_modal}
\begin{figure*}[htp]
\centering

\begin{subfigure}[t]{0.18\textwidth}
  \centering
  \includegraphics[width=0.85\linewidth]{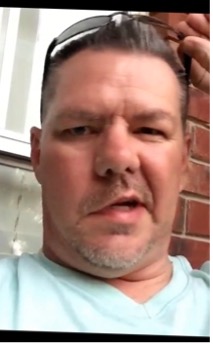}
  \caption{}
  \label{fig:data_3d}
\end{subfigure}
\begin{subfigure}[t]{0.18\textwidth}
  \centering
  \includegraphics[width=0.8\linewidth]{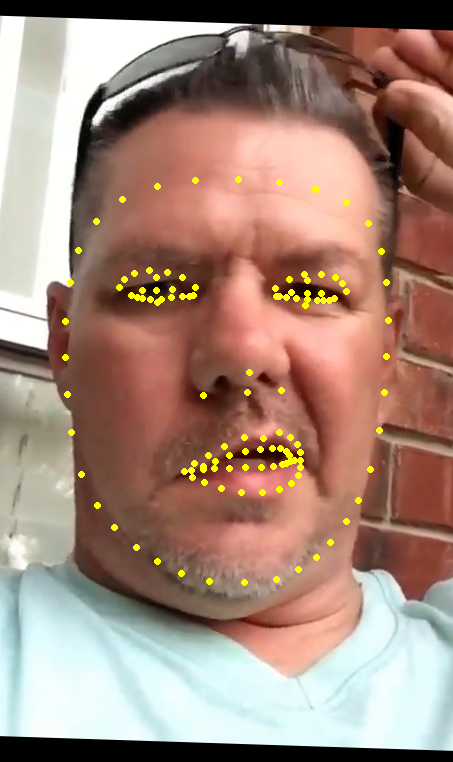}
  \caption{}
  \label{fig:data_3b}
\end{subfigure}
\hspace{1mm}
\begin{subfigure}[t]{0.28\textwidth}
  \centering
  \includegraphics[width=\linewidth]{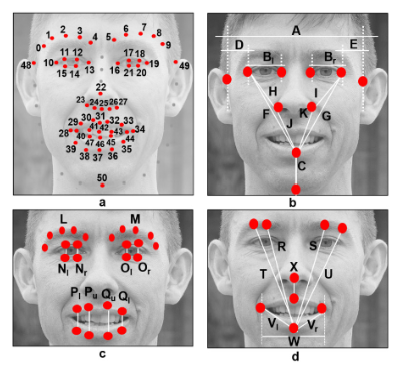}
  \caption{}
  \label{fig:data_3c}
\end{subfigure}
\hspace{1mm}
\begin{subfigure}[t]{0.23\textwidth}
  \centering
  \includegraphics[width=\linewidth]{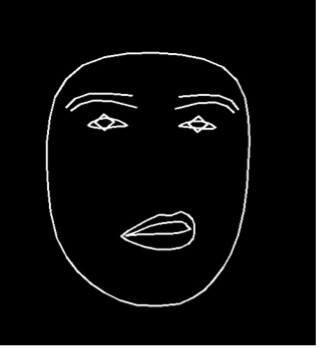}
  \caption{}
  \label{fig:data_3e}
\end{subfigure}

\caption{(a) Raw RGB image, (b) Subset of 478 2-dimensional coordinates of eyes, nose, and mouth regions overlaid on the RGB image, (c) Handcrafted Manual Features, and (d) Black and White (BnW) Line Segments of Facial Contours}
\label{fig:sample_data}
\end{figure*}

To develop an AI model, we processed raw RGB image frames of videos using the facial landmark estimation model (\cite{lugaresi2019mediapipe}) to obtain five data modalities: 1) raw RGB images (Figure \ref{fig:data_3d}), 2) features of facial expressions, 3) facial landmark coordinates (Figure \ref{fig:data_3b}), 4) handcrafted manual features computed using (Figure \ref{fig:data_3c}), and 5) black and white (BnW) images with line segments representing the facial silhouette and local regions (Figure \ref{fig:data_3e}).

For facial landmark coordinates, we include all 478 2-dimensional coordinates of facial landmarks from the model \cite{lugaresi2019mediapipe}. The resulting facial landmark coordinates are a matrix of $\mathbf{m} \in \mathbb{R}^{478\times2}$.

Feature vectors of facial expressions contain 52 distinct attributes, each representing the extent of a facial expression (e.g. the extent to which each eye is closed, or to which the mouth is opened) in a float value within the range [0, 1]. The resulting vector of features of facial expressions is denoted as $\mathbf{b} \in \mathbb{R}^{52}$.

Handcrafted manual features contain 29 distinct attributes, following the methodology outlined by \cite{diagnostics12071528}. Each quantifies the extent of facial asymmetry and expression, with each feature represented as a float value in [0,1]. They characterise the symmetry and alignment of key facial regions, such as the eyebrows, eyes, and mouth. These features capture the relative positions, inclinations, and angles of facial landmarks, as well as ratios that quantify asymmetry between the left and right sides of the face. The resulting vector of features of facial expressions is $\mathbf{h} \in \mathbb{R}^{29}$. 

To obtain the BnW line segment images, we first used the facial landmark estimation model to generate contours of the detected face, eyebrows, and eyes, then plotted them in white against a solid black background (Figure \ref{fig:data_3e}).

\subsection{Model Architectures}
Given five data modalities, we explored different model architectures (i.e. feed-forward neural networks, ResNet50-based \cite{he2016deep}, FasterViT-based \cite{hatamizadeh2023fastervit}, and MLP Mixer-based \cite{tolstikhin2024} models) tailored to each data modality and combined modalities (i.e. early and late fusion).

In contrast to previous studies \cite{hsu2018deep,axioms12121091,oo2024exploring}, our work further explores the effectiveness of patch-based architectures such as MLP Mixer \cite{tolstikhin2024} and FasterViT \cite{hatamizadeh2023fastervit}. These models are structurally similar, as both divide the input image into smaller, non-overlapping patches before using them for computation. By processing patches which are non-overlapping, these models can better focus on facial regions critical for detecting asymmetries associated with facial palsy, ultimately boosting detection accuracy and robustness.

\subsubsection{Problem Formulation}
We formulated the detection of facial palsy as a binary classification task, where the presence of facial palsy is defined as the presence of palsy. This approach allows us to indicate either the presence or absence of facial palsy for each image. 

Given the binary classification task, we applied the Binary Cross-Entropy (BCE) as the objective loss function.

\begin{displaymath}
BCE = -{(y\log(p) + (1 - y)\log(1 - p))}
\end{displaymath}

\subsubsection{Feed-forward Neural Network Model}
Feed-forward Neural Network (FNN) models were trained to perform classification using facial landmark coordinates, feature vectors of facial expressions, and handcrafted manual features separately.


The FNN Model for features of facial expressions comprises three linear layers with intermediate batch normalization and ReLU activations, followed by a sigmoid layer for binary classification. We utilised the hidden unit sizes of 52, 32, 10, 2.

The FNN Model for facial landmark coordinates comprises six hidden layers with dropout regularisation applied at varying probabilities (0.25, 0.3, 0.5, 0.1). We utilised the hidden unit sizes of 956, 512, 256, 128, 64, 32, 2, employing batch normalization and ReLU activations to stabilize and improve learning. The final layer also uses a sigmoid activation for binary classification.

The FNN Model for handcrafted manual features has a consistent width of 59 neurons across its hidden layers. The network consists of multiple linear layers (15 hidden layers total), each followed by ReLU activation and batch normalization. The final layer also uses a sigmoid activation for binary classification.

The FFN architectures and hyperparameters were chosen empirically based on which gave the best performance (e.g. after changing the number of layers, number of neurons). All FNN models were trained with a learning rate of 0.2045, batch size 256, and SGD optimiser. The FFN model for features of facial expressions was trained for 1000 epochs, and the FFN model for coordinates and handcrafted features were trained for 3000 epochs.

\subsubsection{ResNet50-based models}
For ResNet50-based models, we fine-tuned the pre-trained ResNet50 model \cite{he2016deep} to perform classification using raw RGB images or BnW images with line segments. The final layer in the pre-trained ResNet50 model was modified to include a fully-connected layer of 512 hidden units followed by a ReLU activation function, a dropout layer, and a 1-dimensional batch norm layer and another fully-connected layer with a sigmoid activation at the end to output class probabilities. For transfer learning, all model parameters are frozen, except for those in the final layer and modified classification layer. 

The ResNet50-based models were trained using the SGD optimizer with a learning rate of 0.01. For RGB images, we applied a batch size of 256 for 20 epochs. For the black-and white (BnW) line segment images, we applied a batch size of 128 for a maximum of 100 epochs with early stopping based on training loss, and patience of 3 epochs.

\subsubsection{FasterViT-based models}
FasterViT \cite{hatamizadeh2023fastervit} leverages a hybrid convolution-attention design where patches enable parallel processing, reducing memory usage while maintaining high accuracy. For FasterViT-based models, we fine-tuned the pre-trained FasterVit-0 model \cite{hatamizadeh2023fastervit} to perform classification using raw RGB images or BnW images with line segments. For transfer learning, all model parameters are frozen, except for those in the final convolution-attention layer. 

For RGB images, the FasterViT model was trained with 0.001 learning rate, batch size 256 and AdamW optimizer for 20 epochs. For BnW line segment images, the FasterViT-based model was trained using the AdamW optimizer with a learning rate of 0.001, and a batch size of 128 for a maximum of 100 epochs with early stopping based on training loss, and patience of 3 epochs.

\subsubsection{MLP Mixer-based models}
In MLP Mixer \cite{tolstikhin2024}, patch embeddings allow efficient global feature mixing through token-mixing and channel-mixing layers, enabling the capture of both spatial and semantic information. For MLP Mixer-based models, we fine-tuned the pre-trained MLP Mixer B-16 model \cite{tolstikhin2024} to perform classification using raw RGB images or BnW images with line segments. For transfer learning, all model parameters are frozen, except for those in the final MLP-Mixer block and classification head. 

The MLP Mixer-based models were trained with a learning rate of 0.01  and SGD optimizer. For RGB images, the MLP Mixer-based model was trained with a batch size of 256 for 40 epochs. For BnW line segment images, the MLP Mixer-based model was trained with a batch size of 128 for a maximum of 100 epochs with early stopping based on training loss, and patience of 3 epochs.

\subsubsection{Early \& Late Fusion Models}
With the goal of improving facial palsy detection, we explored a multimodal fusion approach \cite{baltruvsaitis2018multimodal} using both structured and unstructured data. Specifically, we utilized model-agnostic approaches: 1) early fusion that integrates two data modalities after extracting their features and concatenating their features, and 2) late fusion that integrates the outputs of the models using each of the modalities (Figure 
\ref{fig:model_architecture}).

For our fusion models, we utilized the models that were independently trained using a single modality (Section \ref{sect:data_modal}) and then empirically identified two modalities that achieved the highest performance in facial palsy detection. 

For early fusion, we extracted the feature vectors from two models that were independently trained using a single modality and concatenated these feature vectors to train an early fusion model (Figure \ref{fig:model_architecture}). We empirically chose the intermediate layer of a model using a single model to extract a feature vector. The early fusion model consists of four fully connected layers, in which we applied batch normalization for regularization with LeakyReLU activation functions.

For late fusion, we computed the average of the output probabilities from two models that were independently trained using a single modality (e.g. handcrafted features or raw RGB images) and generated the model predictions on the class with the highest mean probability. 

Early fusion models were trained using the SGD optimizer with a batch size of 128 for a maximum of 100 epochs with early stopping based on training loss similar to the Bnw line segment image models. The fusion model using handcrafted features + RGB image embedding from the MLP Mixer-based model was trained with a learning rate of 0.01, while the fusion model using the handcrafted features and BnW line segment embedding from the ResNet50-based model was trained with a learning rate of 0.001.

\section{Experiments} \label{section:experiments}
\subsection{Dataset}
Our experiments were conducted using two datasets: the YouTube Facial Palsy (YFP) dataset by \cite{hsu2018deep} and the Extended Cohn-Kanade (CK+) dataset \cite{kanade2000comprehensive,lucey2010extended}.

The YFP dataset has been labeled by three independent clinicians. It consists of 31 videos collected from 21 facial palsy patients, and some patients have provided multiple videos. In each video, the patient talks to the camera, recording variations in their facial expressions over time. Each video has been converted to an image sequence with a sampling rate of 6 FPS. For each image in the sequence, the dataset contains labels indicating the intensity of palsy in two facial regions (i.e. the eyes and mouth) where it can be observed.

The CK+ dataset consists 593 video sequences from 123 different subjects aged 18-50 years, with gender and racial diversity. Each video is recorded at 30 FPS and frames are labelled with one of seven expressions: anger, contempt, disgust, fear, happiness, sadness, and surprise. The CK+ dataset is posed, meaning that all facial expressions are intentionally acted out by the subjects in a controlled environment. The CK+ dataset is included to ensure robustness to variations in facial expressions.

\subsection{Evaluation}
\subsubsection{Data Splitting Strategy}
For the evaluation of different data modalities and model architectures, we applied the leave-one-patient-out (LOPO) cross-validation, in which we trained the model using data except for held-out data from a patient for testing and we repeated this process over 21 patients. For each iteration of LOPO, the training dataset consists of samples from 20 palsy patients from the YFP dataset and 20 healthy subjects from the CK+ dataset. 50 samples are collected from each palsy patient and healthy subject. 

For each iteration of LOPO, the testing dataset consists of samples collected from the remaining held-out palsy patient and another 20 healthy subjects from the CK+ dataset. 50 samples are collected from the held-out palsy patient, and 2 samples are collected from each of the 20 CK+ subjects used for testing. 

For each LOPO iteration, we recorded the F1-score, precision, and recall and then averaged these metrics for each data modality and model architecture.

\subsubsection{Data Sampling Strategy}
Without careful sampling for palsy patients, some levels of palsy severity (e.g., SlightPalsy-Eyes-Normal-Mouth) might be overrepresented, while others might be underrepresented with random sampling for each palsy patient. This issue can happen even if the patient has samples for all label combinations. To address this, while the target sample count for each subject is not reached, we select one sample from each unique palsy intensity label class in a round-robin fashion.

\section{Results \& Discussion} \label{section:results}

Table \ref{tab:results} presents the experimental outcomes across various data modalities alongside the corresponding single-modal or multimodal models trained on them.

\begin{table}[htp]
\centering
\caption{Results of different data modalities and model architectures}
\label{tab:results}
\resizebox{\textwidth}{!}{%
\begin{tabular}{cccccc} \toprule
\textbf{Data Modality} &
  \textbf{Model} &
  \textbf{Average F1} &
  \textbf{Average Precision} &
  \textbf{Average Recall} \\ \midrule
Features of Facial Expressions  & Feed-forward Neural Network      & \textbf{83.95} & \textbf{88.65} & 82.50     \\
Coordinates  & Feed-forward Neural Network      & 67.80          & 76.80          & 66.60      \\
Handcrafted Manual Features \cite{diagnostics12071528}  & Feed-forward Neural Network      & 82.80          & 82.85         & \textbf{88.30}     \\
\midrule

RGB Images   & ResNet50 & 85.00          & 96.00 & 76.70      \\ 
RGB Images   & FasterViT & 84.00          & 89.90 & 81.50      \\ 
RGB Images   & MLP Mixer & \textbf{89.00}          & \textbf{96.60} & \textbf{88.00}      \\ 
\midrule
BnW LineSegment Images       & ResNet50 & \textbf{77.00} & \textbf{85.40} & \textbf{70.70}  \\ 
BnW LineSegment Images       & FasterViT & 67.00 & 83.84 & 58.23  \\ 
BnW LineSegment Images       & MLP Mixer & 58.00 & 78.50 & 49.95  \\ \midrule \midrule

\begin{tabular}[c]{@{}c@{}}Handcrafted Manual Features +\\RGB Images \end{tabular} &
  \begin{tabular}[c]{@{}c@{}}Early Fusion Model\\ (MLP Mixer + Feedforward Model)\end{tabular} &
  \textbf{96.00} &
  \textbf{94.80} &
  \textbf{98.30} &
   \\ \midrule

\begin{tabular}[c]{@{}c@{}}Handcrafted Manual Features +\\BnW LineSegment Images\end{tabular} &
  \begin{tabular}[c]{@{}c@{}}Early Fusion Model\\(ResNet50 + Feedforward Model)\end{tabular} &
  79.90 &
  {94.00} &
  76.50 \\ \midrule
\begin{tabular}[c]{@{}c@{}}Handcrafted Manual Features +\\RGB Images\end{tabular} &

\begin{tabular}[c]{@{}c@{}}Late Fusion Model\\(MLP Mixer + Feedforward Model)\end{tabular} &
  83.00 &
  {77.95} &
  94.70 \\ \midrule
\begin{tabular}[c]{@{}c@{}}Handcrafted Manual Features +\\BnW LineSegment Images\end{tabular} &
  
  \begin{tabular}[c]{@{}c@{}}Late Fusion Model\\(ResNet50 + Feedforward Model)\end{tabular} &
  72.45 &
  {75.60} &
  81.25 \\ \bottomrule 
\end{tabular} 
} 
\end{table}

Among structured data modalities, the \textbf{FFN model using features of facial expressions achieved the highest F1-score of 83.95 and the highest precision of 88.65}. However, the \textbf{FFN model using handcrafted manual features achieved the highest recall of 88.30} for a similar F1-score of 82.80. Given the higher cost associated with missing cases of facial palsy, we prioritize the handcrafted manual features modality to minimize false negatives.

Among the models trained on RGB images (unstructured data), the \textbf{MLP Mixer model achieved the highest F1-score of 89.00, precision of 96.60, and recall of 88.00}. Among the models trained on BnW Line Segment images (unstructured data), {the ResNet50 model achieved the highest F1-score of 77.00, precision of 85.40, and recall of 70.70}.
Given these results, we further explored the benefit of integrating the data modalities of raw RGB images using MLP Mixer-based embeddings with handcrafted manual features.

When combining structured data (handcrafted manual features) and unstructured data (raw RGB images), the \textbf{early fusion model demonstrated the best overall performance, achieving the highest F1-score of 96.00 and recall of 98.30}, significantly outperforming both single-modality models. Specifically, the early fusion model surpassed the performance of the FFN trained on manual features by 13.20 in F1-score and 10.00 in recall respectively and the MLP Mixer-based model trained on RGB images by 7.00 and 10.30 in F1-score and recall respectively. While the MLP Mixer-based model trained on RGB images achieved the highest precision of 96.60, its F1-score of 89.00 and recall of 88.00 were lower compared to the early fusion model.

Interestingly, when utilizing with RGB images, the MLP Mixer-based model (F1-score of 89.00 and precision of 96.60) outperformed the ResNet50-based model (F1-score of 85.00 and precision of 96.00) by achieving higher average F1-score and precision respectively. This suggests that the token-mixing mechanism of the MLP Mixer-based model, which processes input images by dividing them into fixed-size patches and independently applying transformations along spatial and channel dimensions, may offer a competitive alternative to traditional convolutional architectures. Its spatial token mixing enables the model to capture long-range dependencies across patches, while channel mixing enhances feature extraction by processing information across multiple feature dimensions.

Future work will focus on refining fusion techniques and exploring video-based, temporal approaches \cite{lee2024towards} to further enhance detection accuracy and robustness. In addition, we will study how to make an AI output explainable \cite{doshi2017towards,lee2024towards} to clinicians, aiming to support effective clinicial assistance to improve the clinical practices \cite{ghassemi2021false,lee2021human}.

\section{Conclusion} \label{section:conclusion}
We contribute to a comprehensive analysis of deep learning models utilizing different data modalities, including RGB images, facial landmark coordinates, and handcrafted feature vectors and the combinations of unstructured and structured data modalities. Our findings demonstrate that models leveraging single modalities, such as handcrafted features or RGB images, performed well individually. The feed-forward neural network using handcrafted features achieved an F1 score of 82.80 while the MLP Mixer-based model using RGB images reached an F1 score of 89.00. Moreover, our proposed multimodal fusion-based deep learning model, which integrates RGB images with the MLP Mixer-based model and handcrafted features, significantly improved the F1 score to 96.00, highlighting the benefits of combining diverse data modalities. By utilizing the publicly available YouTube Facial Palsy dataset and the CK+ emotion dataset, and providing a detailed comparison of performance across various deep learning models, this study provides a solid foundation for future benchmarking efforts.

\begin{credits}
\subsubsection{\ackname} This research is supported by the National Research Foundation, Singapore
under its AI Singapore Programme (AISG Award No: AISG3-AMP-2024-08-003) and partially supported by the Ministry of Education, Singapore under its Academic Research Fund Tier 2 (MOE-T2EP20223-0007). Any opinions, findings and conclusions or recommendations expressed in this material are those of the authors and do not reflect the views of the Ministry of Education, Singapore.
\end{credits}

%
%
%
\bibliographystyle{splncs04}
\bibliography{main}
%
\end{document}